\documentclass[10pt,twocolumn,letterpaper]{article}

\usepackage{iccv}
\usepackage{times}
\usepackage{epsfig}
\usepackage{graphicx}
\usepackage{amsmath}
\usepackage{amssymb}

\usepackage{epsfig}
\usepackage{graphicx}
\usepackage{amsmath}
\usepackage{amssymb}
\usepackage{multirow}
\usepackage{caption}
\usepackage{subcaption}
\usepackage[skip=0.5\baselineskip]{caption}
\usepackage{booktabs}
\usepackage{ textcomp }
\PassOptionsToPackage{dvipsnames}{xcolor}
\usepackage[dvipsnames]{xcolor}
\newcommand{\splitcell}[1]{\begin{tabular}{@{}c@{}}#1\end{tabular}}
\newcommand{\bsplitcell}[1]{$\left[\splitcell{#1}\right]$}

\usepackage[pagebackref=true,breaklinks=true,letterpaper=true,colorlinks,bookmarks=false]{hyperref}

\iccvfinalcopy 

\def\iccvPaperID{10078} 

\begin{document}

\title{Diverse Temporal Aggregation and Depthwise Spatiotemporal Factorization for Efficient Video Classification}

\title{Diverse Temporal Aggregation and Depthwise Spatiotemporal Factorization for Efficient Video Classification}

\author{%
      Youngwan Lee\ \ \ Hyung-Il Kim\ \ \ Kimin Yun\ \ \ Jinyoung Moon\\
  Electronics and Telecommunications Research Institute (ETRI), South Korea\\
  \tt\small\texttt{\textbraceleft yw.lee, hikim, kimin.yun, jymoon\textbraceright@etri.re.kr} \\
}
\maketitle

\begin{abstract}

Video classification researches have recently attracted attention in the fields of temporal modeling and  efficient 3D architectures.
However, the temporal modeling methods are not efficient, and there is little interest in how to deal with temporal modeling in a 3D efficient architecture.
For bridging the gap between them, we propose an efficient 3D architecture for temporal modeling, called VoV3D, that consists of a temporal one-shot aggregation (T-OSA) module and depthwise factorized component, D(2+1)D.
The T-OSA is devised to build a feature hierarchy by aggregating spatiotemporal features with different temporal receptive fields.
Stacking this T-OSA enables the network itself to model short-range as well as long-range temporal relationships across frames without any external modules.
We also design a depthwise spatiotemporal factorization module, named, D(2+1)D that decomposes a 3D depthwise convolution into two spatial and temporal depthwise convolutions for making our network more lightweight and efficient.
Through the proposed temporal modeling method (T-OSA) and the efficient factorized module (D(2+1)D), we construct two types of VoV3D networks, VoV3D-M and VoV3D-L.
Thanks to its efficiency and effectiveness of temporal modeling, VoV3D-L has $6\times$ fewer model parameters and $14\times$ less computation, surpassing state-of-the-arts on both Something-Something and Kinetics-400 datasets.
Furthermore, VoV3D shows better abilities for temporal modeling than state-of-the-art efficient 3D architectures.
We hope that VoV3D can serve as a baseline for efficient temporal modeling architecture.
\end{abstract}


\section{Introduction}
Recently, many works~\cite{lin2019tsm,jiang2019stm,tan2018mnasnet,yang2020temporal,fan2020pyslowfast,wang2016temporal} for video classification have been studied to handle temporal modeling that characterizes the temporal variation and dynamics of an action (\textit{i.e.}, \textit{visual tempo}~\cite{yang2020temporal}).
Unlike 2D image classification, video classification should distinguish visual tempo variation as well as its semantic appearance.
In other words, appearance information alone is not sufficient to distinguish between \textit{moving something up and down} or between \textit{walking and running}, which requires to capture visual tempo variations.
Thus, effectively modeling visual tempo is a key factor for video classification.

Previous works~\cite{wang2016temporal,jiang2019stm,lin2019tsm,li2020tea} for temporal modeling utilize 2D CNN architecture due to its efficiency rather than 3D CNN one, which usually process per-frame inputs and aggregate these results to produce a final output by adopting temporal shift module~\cite{lin2019tsm} or motion information embedding module~\cite{jiang2019stm,tan2018mnasnet}.
However, these methods depend heavily on the 2D ResNet~\cite{he2016deep} backbone, which is neither lightweight nor efficient compared to state-of-the-art efficient 2D CNN models~\cite{tan2018mnasnet,howard2019searching,tan2019efficientnet}.
3D CNN-based temporal modeling methods~\cite{fan2020pyslowfast,yang2020temporal} are also proposed to construct input frame-level pyramid~\cite{feichtenhofer2019slowfast} with different input frame rates or feature-level pyramid~\cite{yang2020temporal} with dynamic visual tempo modeling.
However, these methods require extra model capacity by adding a separate network path or a fusion module.
In short, since previous works are add-on style modules on top of the backbone network, they are constrained under the backbone.

Another research that has recently attracted attention for video understanding is to build efficient network architectures~\cite{kopuklu2019resource,tran2019video,feichtenhofer2020x3d}.
These works exploit 3D depthwise convolution for reducing model parameters and computations like 2D efficient CNN architectures~\cite{howard2017mobilenets,sandler2018mobilenetv2,howard2019searching,tan2018mnasnet,ma2018shufflenet,tan2019efficientnet} that replace a convolution with a combination of depthwise convolution and pointwise convolution.
This depthwise separable convolution can be called channel factorization.
However, these 3D efficient networks only focus on building architectures and do not consider temporal modeling.

\begin{figure}[t]
\centering
  \scalebox{0.54}{
  \includegraphics{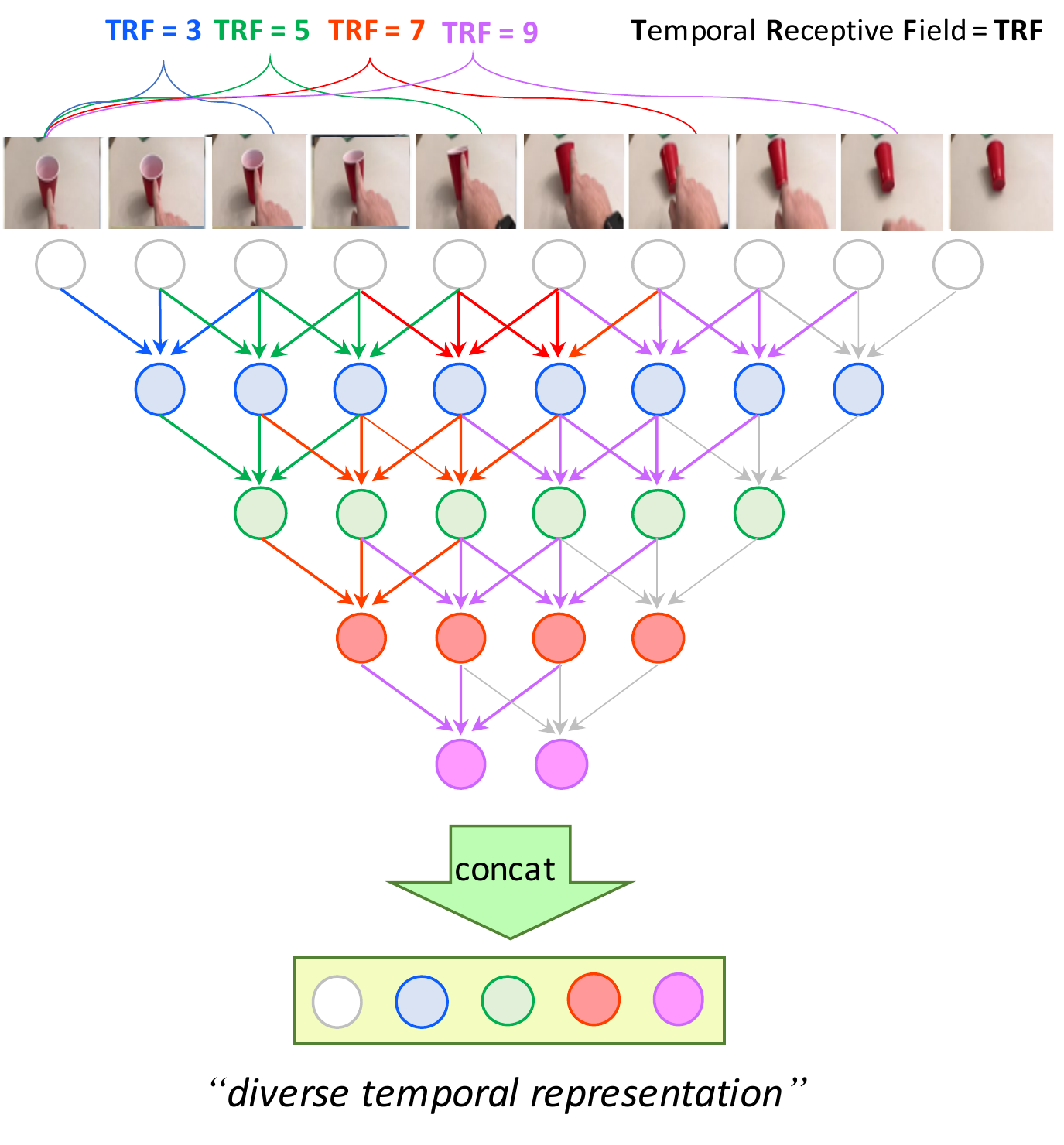} 
  }
  \caption{\textbf{Illustration of the temporal receptive field.} The features having multiple temporal receptive fields are advantageous to capture visual tempo variation of an action.    
  } 
\label{fig:trf}
\end{figure}

For addressing these issues, in this work, we propose an efficient and effective temporal modeling 3D architecture, called VoV3D.
The proposed VoV3D consists of temporal one-shot aggregation (T-OSA) building blocks, which  are made of the proposed depthwise factorized module (\textit{i.e.}, D(2+1)D).
The T-OSA is devised to build a temporal feature hierarchy by aggregating features with different temporal receptive fields.
As illustrated in Fig.~\ref{fig:trf}, having multiple temporal receptive fields is helpful to capture the visual tempo variation of an action.
From this perspective, stacking the T-OSAs enables the network itself to model short-range as well as long-range temporal relationships across frames without any external modules~\cite{fan2020pyslowfast,yang2020temporal}.

Inspired by the optimization benefit from kernel factorization~\cite{Tran_2018_CVPR,xie2018rethinking} and the efficiency of channel factorization~\cite{tran2019video,feichtenhofer2020x3d}, we also design a depthwise spatiotemporal factorized module, called D(2+1)D that decomposes 3D depthwise convolution into \textit{spatial} depthwise convolution and \textit{temporal} depthwise convolution for making our network more lightweight and efficient.
In practice, we have confirmed that combining the two factorization methods achieves better performance and efficiency than each one. 
Moreover, the efficiency of D(2+1)D makes our network possible to use more input frames (over 16 frames), which is advantageous for temporal modeling.

By using the proposed temporal modeling method, T-OSA, and the efficient factorized module, D(2+1)D, we construct two types of 3D CNN architectures, VoV3D-M and VoV3D-L models.
In order to evaluate the proposed method in terms of modeling temporal variations, we validate VoV3D on the Something-Something dataset~\cite{goyal2017something} which has been well-known to be challenging to classify an action due to the temporal complexity~\cite{zhou2018temporal,xie2018rethinking,lin2019tsm}.
Moreover, we show the performance on Kinetics-400 dataset~\cite{kay2017kinetics} to compare the proposed network to the many state-of-the-arts.
Thanks to its efficiency and effectiveness of the proposed temporal modeling mechanism, VoV3D-M outperforms the state-of-the-art both 2D~\cite{li2020tea} and 3D~\cite{feichtenhofer2019slowfast} temporal modeling methods, while having much $7\times$ and $11\times$ fewer parameters and $27\times$ and $10\times$ less computation on Something-Something dataset~\cite{goyal2017something}.
Furthermore, the proposed VoV3D shows better temporal modeling ability than the state-of-the-art efficient 3D architecture, X3D~\cite{feichtenhofer2020x3d} having comparable model capacity.
We hope that the ideas contained within the proposed VoV3D are widely used in other methods.

The main contributions of this work are summarized as below:
\begin{itemize}
\item We propose an effective temporal modeling method, Temporal One-Shot Aggregation~(T-OSA) that can capture visual tempo variations by aggregating features having different temporal receptive fields. 
\item We propose an efficient depthwise factorized module, D(2+1)D that decomposes a 3D convolution into spatial and temporal depthwise convolutions, making T-OSA modules better accuracy and efficiency.
\item We design an efficient and effective 3D CNN architecture, VoV3D based on the proposed T-OSA and D(2+1)D modules, which outperforms the state-of-the-arts in terms of both temporal modeling and efficiency. 
\end{itemize}

\begin{figure*}[h]
\centering
  \scalebox{0.9}{
  \includegraphics[width=\textwidth]{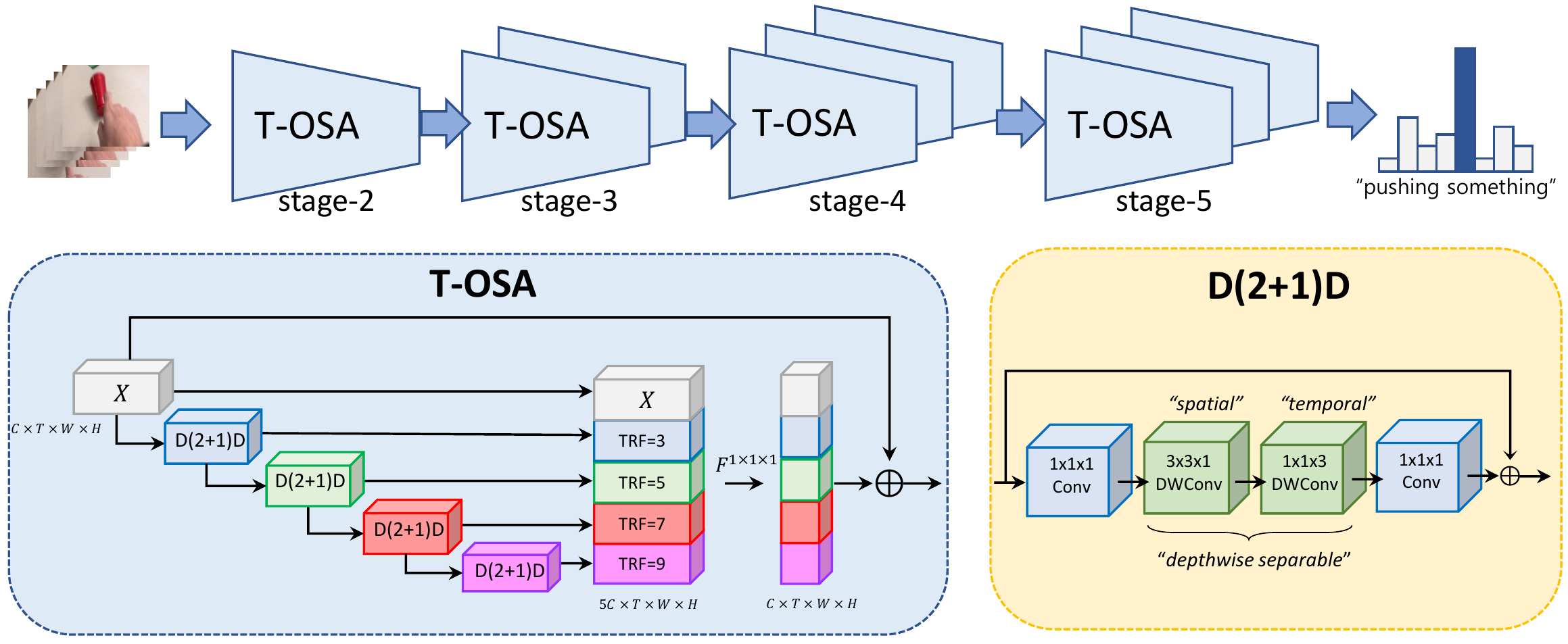} 
  }
  \caption{
  \textbf{VoV3D} has Temporal One-Shot Aggregation~(T-OSA) building blocks. T-OSA consists of depthwise spatiotemporal factorized modules, D(2+1)D.
  Please refer to the details of the VoV3D architecture in Table 2.
  }
  \vspace{-0.2cm}
\label{fig:tosa}
\end{figure*}

\section{Related works}

\subsection{Temporal modeling for video classification}
Recent attempts for temporal modeling for video classification could be divided into two categories: 2D CNN-based and 3D CNN-based methods.
2D CNN-based methods~such as TSN~\cite{wang2016temporal}, TSM~\cite{lin2019tsm}, STM~\cite{jiang2019stm} and TEA~\cite{li2020tea} prefer to use 2D CNN, \eg, ResNet-50 as a backbone, due to its efficiency than 3D CNN models.
They process per-frame inputs and aggregate these results to produce a final output on top of 2D ResNet.
TSN~\cite{wang2016temporal} proposes to form a clip by sampling evenly from divided segments and this sparse sampling method becomes a common strategy for many works.
TSM~\cite{lin2019tsm} is proposed to model temporal motion by utilizing memory shift operation along the temporal dimension.
Since motion information is also an important cue for temporal modeling as a short-term temporal relationship, attempts to model feature-level motion features are proposed in STM~\cite{jiang2019stm} and TEA~\cite{li2020tea}.
STM and TEA propose to differentiate between adjacent features for representing motion features and then add the spatiotemporal features and motion encoding together.
TEA~\cite{li2020tea} also has a temporal aggregation module to capture long-range temporal dependency.
However, TEA is based on 2D CNN features that are not jointly convolved along with spatial and temporal axis.
This means that the interaction between spatial and temporal features in~\cite{li2020tea} is limited than 3D spatiotemporal methods.

For modeling various visual tempos using spatiotemporal 3D CNN, many works have been proposed by building an input frame-level pyramid~\cite{feichtenhofer2019slowfast,zhang2018dynamic} or feature-level pyramid~\cite{yang2020temporal}.
SlowFast~\cite{feichtenhofer2019slowfast} has two network inputs with different frame rates to capture different types of visual information, \textit{e.g.}, semantic appearance or motion.
DTPN~\cite{zhang2018dynamic} also uses a different sampling rate for arbitrary-length input video, which builds up the input frame-level hierarchy.
Unlike these methods, TPN~\cite{yang2020temporal} leverages the feature hierarchy on top of the backbone network, instead of the input frame-level hierarchy by building a temporal feature pyramid network.
In short, since temporal modeling methods are based on the existing backbone networks, \textit{e.g.}, ResNet-50, they are constrained under the nature of the backbone network.

\subsection{Efficient 3D CNN architecture}
Since channelwise separable convolution is densely exploited by efficient 2D CNN architectures~\cite{howard2017mobilenets,sandler2018mobilenetv2,howard2019searching,ma2018shufflenet,ma2018shufflenet,tan2019efficientnet,tan2018mnasnet}, 3D CNN architectures~\cite{tran2019video,feichtenhofer2020x3d,kopuklu2019resource} based on the extended depthwise separable convolution have been explored. 
CSN~\cite{tran2019video} adopts 3D depthwise separable convolution into the residual bottleneck block~\cite{he2016deep} by replacing the $3\times 3\times 3$ convolution and adding a $1\times 1\times 1$ convolution in front of the 3D depthwise convolution for interaction between channels.
X3D~\cite{feichtenhofer2020x3d} explores 3D CNN architecture along with spatial, temporal, depth, channel axis for maximizing the efficacy of the 3D CNN model.
The depthwise bottleneck is also utilized as a key component in X3D, while X3D is progressively expanded from a lightweight to a large-scale model by scale-up all kinds of axes.
As a result, X3D achieves state-of-the-art performance with a much smaller model capacity on various video classification datasets such as Kinetics-400.
However, this method focuses on building an efficient network while temporal modeling is not considered deeply.
Therefore, we focus on building an efficient 3D CNN architecture as well as temporal modeling simultaneously.


\section{VoV3D}
Temporal modeling~(\textit{i.e.}, capturing visual tempo variation) plays an important role in action recognition~\cite{lin2019tsm,jiang2019stm,li2020tea,feichtenhofer2019slowfast,yang2020temporal}.
In particular, in the case of a video that lacks semantic variations of the features, video classification networks should rely heavily on visual tempo.
Moreover, it is necessary to model long-term as well as short-term temporal relationship because short-term information is not sufficient to distinguish visual tempo variations such as \textit{walking vs. running}.
The conventional temporal modeling methods based on 3D CNN try to model the visual tempo through the input frame-level~\cite{zhang2018dynamic,feichtenhofer2019slowfast} or feature-level pyramids~\cite{yang2020temporal}.
However, these methods have to add separate networks on top of the existing 3D backbone as an external~(\textit{i.e.,} plug-in) module, which requires more parameters and computations.
To address these challenges, in this paper, we aim to propose a lightweight and efficient video backbone network having temporal modeling ability by itself without external modules.
To this end, we design a new 3D CNN architecture inspired by VoVNet~\cite{lee2019energy,lee2019centermask} that represents hierarchical and diverse spatial features at a small cost.

First, we propose an effective temporal modeling method, named Temporal One-Shot Aggregation~(T-OSA) inspired by the OSA module in VoVNet~\cite{lee2019energy,lee2019centermask}.
For making a network lightweight and efficient, we also devise a depthwise spatiotemporal factorization component, D(2+1)D.
Lastly, we design a new video classification network, called VoV3D, which is comprised of the proposed T-OSA and D(2+1)D.

\subsection{Temporal One-Shot Aggregation (T-OSA)}


VoVNet~\cite{lee2019energy,lee2019centermask} is a computation and energy-efficient 2D CNN architecture devised to learn diverse feature representations by stacking One-Shot-Aggregation~(OSA) modules.
The OSA module consists of successive $3\times3$ convolutions and aggregates those feature maps into one feature map at once in a concatenate manner, followed by a $1\times1$ convolution.
The OSA allows the network to represent diverse features by capturing multiple receptive fields in one feature map, which results in the effect of the feature pyramid.
Due to the diverse feature representation power of OSA, VoVNet outperforms ResNet~\cite{he2016deep} and HRNet~\cite{sun2019high} in object detection and segmentation tasks that require more complex representation.

Inspired by the spatial feature's hierarchy of OSA in VoVNet, we propose temporal one-shot aggregation, called T-OSA, to capture multiple temporal receptive fields in one 4D feature map, as illustrated in Fig.~\ref{fig:tosa}.
In detail, the $i$-th affine transform $F^{3\times3}_i$~(\eg, $3\times3$ \texttt{2DConv}) can be replaced with $F^{t\times3\times3}_i$~(\eg, $t\times 3\times 3$ \texttt{3DConv}) for $i \in \{1,2, ..., n\}$ where $t$ is the temporal kernel size~(we set to 3) and $n$ is the number of $t\times 3\times 3$ 3D convolutions in T-OSA.
It is noted that we keep temporal dimension $T$~(frames) for feature aggregation.
Each feature map $\mathrm{X}_i \in \mathbb{R}^{C \times T \times W \times H}$ that is the result from $F^{t\times3\times3}_i$ has progressively increasing temporal receptive field due to its successive connection.
For example, if the temporal receptive field~(TRF) of the feature map $\mathrm{X}_1$ is $3$ and temporal kernel size $t$ is 3, the TRF of the next $\mathrm{X}_2$ is $5$.
Thus, once the features are concatenated in channel-axis, the aggregated feature map $\mathrm{X}_{agg} \in \mathbb{R}^{(n+1)C \times T \times W \times H}$ comprised of $\{\mathrm{X}_{in},\mathrm{X}_1,...,\mathrm{X}_{n}\}$ has diverse temporal and spatial receptive fields in one feature map, where $\mathrm{X}_{in}\in \mathbb{R}^{C \times T \times W \times H}$ is the input feature and $n$ is set to 4 in Fig.~\ref{fig:tosa}.
Then, a $1\times1\times1$ convolution is followed for reducing channel size $(n+1)C$ to $C$ and the residual connection is added to the final feature map.
Therefore, stacking T-OSA makes it possible to model short-range as well as long-range temporal dependency across frames, which has a similar effect with feature pyramid in the same spatial feature space.

In practice, simply expanding 2D VoVNet to 3D CNN architecture is limited in terms of optimization because 3D CNN models have additional parameter space along with temporal-axis and thus need optimization strategy.
Therefore, we elaborate the T-OSA with additional design choices for adaptation of OSA in 3D temporal feature space.
While the OSA module in 2D VoVNet uses only $3\times3$ 2D \texttt{Conv} as an affine tranform, the proposed T-OSA adopts 3D bottleneck architecture~\cite{hara2018can, feichtenhofer2020x3d,tran2019video}~(\eg, $1\times 1\times 1$ \texttt{3DConv}, $3\times 3\times 3$ \texttt{3DConv}, $1\times 1\times 1$ \texttt{3DConv}) with more non-linearity operations.
As a 3D bottleneck architecture, we propose D(2+1)D in the next section.
Also, we add an inner residual connection to facilitate optimization.


\begin{figure}[t]
\centering
  \scalebox{0.35}{
  \includegraphics{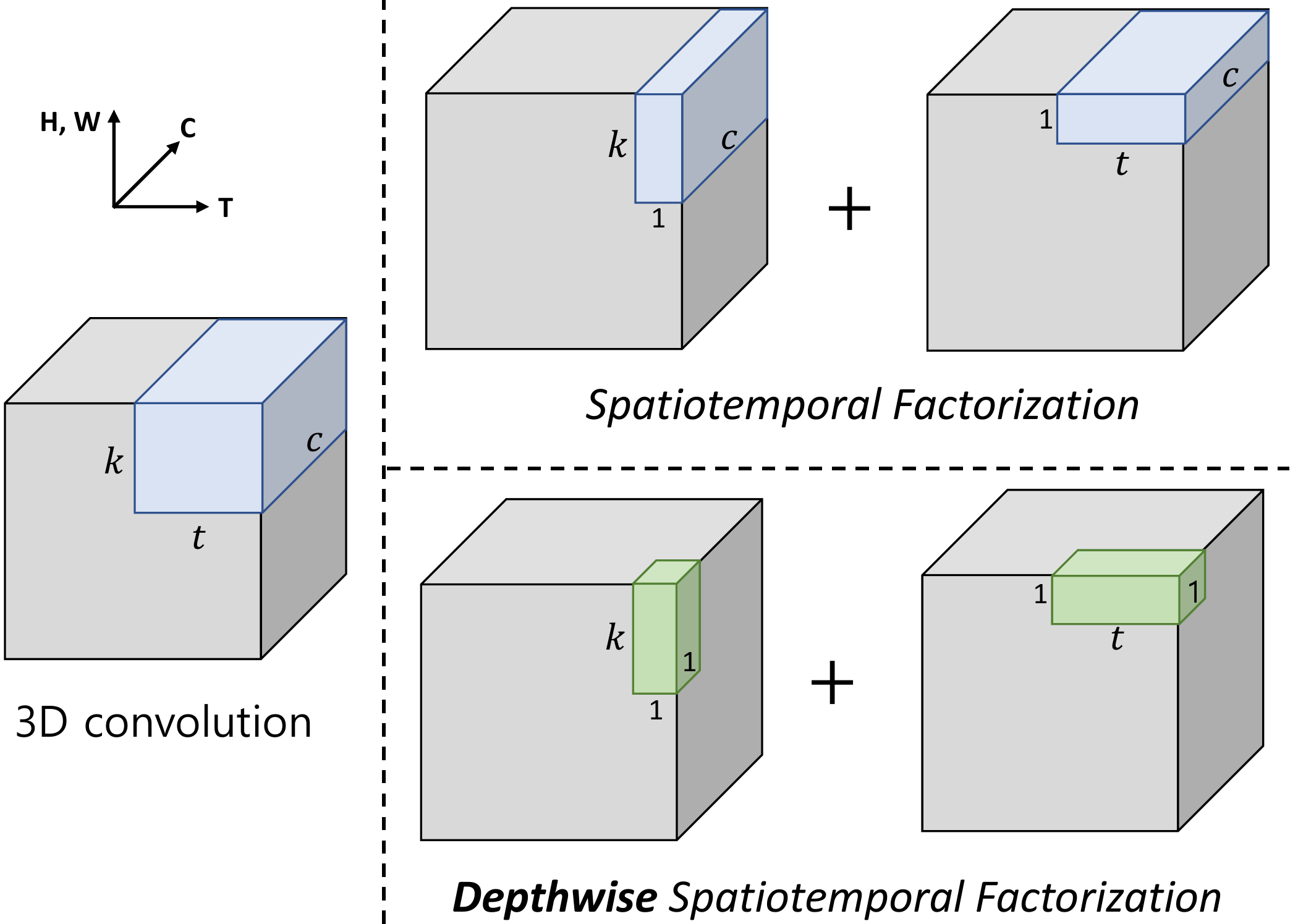} 
  }
  \caption{
  \textbf{Depthwise spatiotemporal factorization.} $k, t, c$ denote the size of spatial kernel, temporal kernel, and channel in 3D convolution, respectively. 
Compared to Spatiotemporal factorization~(top) as in R(2+1)D~\cite{Tran_2018_CVPR}, depthwise spatiotemporal factorization~(bottom) in the proposed D(2+1)D further decomposes the features along the channel axis, which improves the efficiency of the network. 
  }
\label{fig:d21d}
\end{figure}

\subsection{Depthwise Sptaiotemporal Factorization}\label{sec:d2d}
There are two types of factorization concept on 3D convolution~(\texttt{3DConv}): 1) Depthwise ~(or Channelwise)~\cite{tran2019video,feichtenhofer2020x3d,kopuklu2019resource} and 2) Kernelwise~\cite{Tran_2018_CVPR,qiu2017learning,xie2018rethinking} methods.
Inspired by efficient 2D image classification network~\cite{howard2017mobilenets,sandler2018mobilenetv2,howard2019searching,Zhang_2018Shufflenet,ma2018shufflenet,tan2018mnasnet,tan2019efficientnet}, depthwise separable convolution is also mainly used as a key building block for efficient video backbone networks~\cite{tran2019video,feichtenhofer2020x3d,kopuklu2019resource}.
3D depthwise separable convolution~(\texttt{3DWConv}) is utilized to factorize a \texttt{3DConv} into a $t\times k\times k$ depthwise \texttt{3DConv} followed by $1\times1\times1$ pointwise \texttt{3DConv}.
CSN~\cite{tran2019video} adds a $1\times1\times1$ \texttt{3DConv} in front of the \texttt{3DWConv} for preserving the interaction between channels, which results in improving accuracy.
Tran~\etal~\cite{tran2019video} found that the \texttt{3DWConv} has two advantages: 1) significant reduction of parameters and computational cost~(FLOPs) without sacrificing accuracy, 2) regularization effect.
In addition to channel factorization, kernel factorization also has been widely used in ~\cite{Tran_2018_CVPR,qiu2017learning,xie2018rethinking} for curtailing computation and boosting accuracy.
The kernel factorization is also called spatiotemporal factorization as it is decomposed into a $1\times k\times k$ spatial convolution~(space) followed by a $t\times1\times1$ temporal convolution~(time) as shown in Fig.~\ref{fig:d21d}~(top).

\begin{table}[t]
\scalebox{0.94}{
    \begin{tabular}{llr}
    \toprule
    Type & \multicolumn{1}{c}{Param.} & \multicolumn{1}{c}{FLOPs}  \\
    \midrule
    (a) bottleneck   & $C^2tk^2$  & $C^2tk^2(HWT)/s^2$ \\
    (b) R(2+1)D   & $C^2(t+k^2)$   & $C^2(t+k^2)(HWT)/s^2$ \\
    (c) dw-bottleneck   & $Ctk^2$   & $Ctk^2(HWT)/s^2$ \\
    (d) D(1+2)D   & $C(t+k^2)$  & $C(s^2t+k^2)(HWT)/s^2$ \\
    (e) D(2+1)D   & $C(t+k^2)$  & $C(t+k^2)(HWT)/s^2$ \\
    \bottomrule
    \end{tabular}%
}
  \caption{\textbf{Comparison of parameters and computation.} This table considers only a 3D convolution located in the middle of the bottleneck. $t$, $k$, and $s$ denote temporal, spatial kernel size, and stride, respectively. $C$, $H$, $W$, $T$ denote channel, height, width, the number of frames in the input 3D feature map, assuming input/output channel size is same.}
  \label{tab:spec}%
  \vspace{-0.3cm}
\end{table}%

Our motivation lies in the fusion of these two factorization methods for realizing an efficient video classification network.
We design a depthwise spatiotemporal factorized module, \textit{D(2+1)D}, that decomposes a \texttt{3DWConv} into a spatial \texttt{DWConv} and a temporal \texttt{DWConv} as shown in Fig.~\ref{fig:d21d}~(bottom).
We analyze each resource requirement of models in Table~\ref{tab:spec} illustrating the number of parameters and computation~(FLOPs) of a \texttt{3DConv} in the middle of bottleneck architecture.
The input tensor of the \texttt{3DConv} has $C\times T\times H\times W$ shape, where $T$ and $C$ are the number of frames and channels, and $H, W$ is the size of height and width, respectively.
Assuming the number of filters (output channel) is the same~($C$), the 3D filter has $t\times k \times k$ kernel size, where $t, k$ denote temporal and spatial kernel, respectively.
As demonstrated in Table~\ref{tab:spec}, compared to the basic bottleneck \texttt{3DConv} in~(a), \texttt{3DWConv} in~(c) is $C\times$ more efficient because it has only one sub-filter for the input tensor as illustrated in Fig.~\ref{fig:d21d}.
We design two types of factorized modules based on the order of spatial and temporal dimensions: D(1+2)D and D(2+1)D.
It is noted that spatial down-sampling is operated in the spatial convolution and the temporal convolution keeps temporal dimension.
Compared with \texttt{3DWConv} in (c), both D(1+2)D and D(2+1)D have about one order of magnitude fewer parameters and computations.
In comparison between the two factorized modules, an important difference arises in spatial down-sampling.
The number of parameters is the same, while the computation cost is different due to different spatial sizes.
Specifically, for D(1+2)D, the temporal \texttt{DWConv} is operated first with $C\times T\times H\times W$ input tensor followed by the spatial \texttt{DWConv} with stride $s$. 
It is summarized as:
\begin{equation}
\begin{split}
\mathrm{FLOPs} &= Ct\times HWT + Ck^2\times HWT/{s^2} \\
      &= (s^2t+k^2)CHWT/{s^2}.
\end{split}
\end{equation}   
For D(2+1)D, since spatial \texttt{DWConv} with down-sampling goes ahead, the temporal \texttt{DWConv} operates the spatially down-sized input tensor, which results in reducing overall computation.
This is summarized as:
\begin{equation}
\begin{split}
\mathrm{FLOPs} &= Ct\times HWT/{s^2} + Ck^2\times HWT/{s^2}\\
      &= (t+k^2)CHWT/{s^2}.
\end{split}
\end{equation}

It is worth noting that the proposed D(2+1)D shows better efficiency than kernel or channel factorization alone.
We also expect that the D(2+1)D can be widely used for other 3D CNN architectures to boost their performances.
We also have confirmed the effect through the combination of the state-of-the-art method~(\textit{i.e.}, X3D~\cite{feichtenhofer2020x3d}) and our D(2+1)D, which will be described in the experimental section.


\begin{table}[t]
\small
\scalebox{0.88}{
  \begin{tabular}{c|c|c}
  \toprule
  Stage   & VoV3D-M~(\textcolor{blue}{L}) & output size {\small$T\times H\times W$}  \\ \hline
  conv1   & $1\times3^2, 5\times1^2, 24$ & $T\times112\times112$  \\ \hline
  T-OSA2  & \bsplitcell{(D(2+1)D, 40(\textcolor{blue}{48}))$\times5 $ \\ $1\times1^2$, 24}$\times1 (\textcolor{blue}{1})$ & $T\times56\times56$   \\ \hline 
  T-OSA3  & \bsplitcell{(D(2+1)D, 80(\textcolor{blue}{96}))$\times5$ \\ $1\times1^2$, 48}$\times1 (\textcolor{blue}{2})$ & $T\times28\times28$   \\ \hline
  T-OSA4  & \bsplitcell{(D(2+1)D, 160(\textcolor{blue}{192}))$\times5$ \\ $1\times1^2$, 96}$\times2 (\textcolor{blue}{5})$ & $T\times14\times14$  \\ \hline
  T-OSA5  & \bsplitcell{(D(2+1)D, 320(\textcolor{blue}{384}))$\times5$ \\ $1\times1^2$, 160(\textcolor{blue}{192})}$\times2 (\textcolor{blue}{3})$ & $T\times7\times7$ \\ \hline
  conv5   & $1\times1^2, 320(\textcolor{blue}{384})$ & $T\times7\times7$  \\ 
  pool5   & $T\times7\times7$ & $1\times1\times1$  \\ 
  fc1     & $1\times1^2, 2048$ & $1\times1\times1$ \\ 
  fc2     & $1\times1^2$ \#classes & $1\times1\times1$ \\
  \bottomrule
  \end{tabular}
}
  \caption{\textbf{VoV3D architectures: VoV3D-M and VoV3D-L.} $T$ denotes the number of input frames. VoV3D has two types of models: VoV3D-M and VoV3D-\textcolor{blue}{L}. They are comprised of Temporal One-Shot Aggregation (T-OSA) building blocks made of D(2+1)D modules. 
  }
  \label{tab:vov3d}%
  \vspace{-0.3cm}
\end{table}

\subsection{VoV3D Architecture}

Finally, we construct a lightweight and efficient 3D CNN architecture, VoV3D, that can model various visual tempos effectively with the proposed T-OSA and D(2+1)D modules.
We design two types of lightweight models: VoV3D-M and VoV3D-L which have only 3.3M and 5.8M parameters, respectively.
VoV3D is comprised of the proposed T-OSA blocks which consist of five D(2+1)D modules followed by a $1\times1\times1$ \texttt{3DConv}.
In stage level~(same spatial resolution), VoV3D has multiple T-OSAs~(\eg, 5), in series, which leads to representing diverse temporal features.
\texttt{conv1} is also the (2+1)D style-convolution where $1\times3^2$ spatial \texttt{3DConv} is operated and followed by a $3\times1^2$ temporal \texttt{3DConv}.
Following ~\cite{feichtenhofer2020x3d}, we also add a channel attention module, SE block~\cite{hu2018squeeze}, into the D(2+1)D with reduction ratio of 1/16. 
The lightweight and efficient D(2+1)D allows VoV3D to reduce significant computation cost, so it can use longer frames~($\geq $16) to capture longer visual tempo.
The details are illustrated in Table.~\ref{tab:vov3d}.


\section{Experiments}


\begin{table}[t]
\begin{tabular}{@{}lccll@{}}
\toprule
Model        & T-OSA      & D(2+1)D    & Top-1                & Top-5                 \\ \midrule
Baseline (M) &            &            & 46.4             & 75.3                  \\
             & \checkmark &            & 48.0 \scriptsize{+2.4} & 76.7 \scriptsize{+1.4}  \\
             &            & \checkmark & 48.5 \scriptsize{+2.3} & 76.9 \scriptsize{+1.6}  \\
             & \checkmark & \checkmark & 49.0 \scriptsize{+2.6} & 78.2 \scriptsize{+2.9}  \\ \midrule
Baseline (L) &            &            & 47.1                 & 76.5                      \\
             & \checkmark &            & 48.9 \scriptsize{+1.8} & 77.6 \scriptsize{+1.1}  \\
             &            & \checkmark & 48.8 \scriptsize{+1.7} & 77.4 \scriptsize{+0.9}  \\
             & \checkmark & \checkmark & 49.6 \scriptsize{+2.5} & 78.1 \scriptsize{+1.6}  \\ \bottomrule
\end{tabular}
\caption{\textbf{Contributions of the proposed components in VoV3D} on Something-Something V1.}
  \label{tab:ablation}%
\vspace{-0.3cm}
\end{table}

\subsection{Datasets}
We validate the proposed VoV3D on Something-Something (V1 \& V2)~\cite{goyal2017something} and Kinetics-400~\cite{kay2017kinetics}.
In contrast to Kinetics-400~\cite{kay2017kinetics} that is less sensitive to visual tempo variations, Something-Something~\cite{goyal2017something} is focused on human-object interaction which requires a more temporal relationship than appearance~\cite{zhou2018temporal,xie2018rethinking,lin2019tsm}.
Since Something-Something is widely used as a benchmark for evaluating the effectiveness of temporal modeling, the effectiveness of the proposed VoV3D is mainly investigated for this dataset.
Something-Something V1~\cite{goyal2017something} contains 108k videos with 174 categories, and the second release (V2) of the dataset is increased to 220k videos.
Kinetics-400 dataset~\cite{kay2017kinetics} includes 400 categories and provides download URL links over $240$k training and $20$k validation videos.
Because of the expirations of some YouTube links, we collect 234,619 training and 19,761 validation videos.
For the fair comparison with X3D, we train X3D and VoV3D on the same collected Kinetics-400 dataset by ourselves.

\begin{table}[t]
  \scalebox{0.87}{
    \begin{tabular}{lcccc}
    \toprule
    VoV3D-M & Param. & GFLOPs & Top-1  & Top-5 \\
    \midrule
    (a) bottleneck~\cite{hara2018can} & 42.9M & $103.2\times6$  & 48.6   & 76.8 \\
    (b) R(2+1)D~\cite{Tran_2018_CVPR} & 20.9M  & $48.9\times6$    & 48.6   & 77.6 \\
    (c) dw-bottleneck~\cite{feichtenhofer2020x3d,tran2019video} & 3.3M  & $7.0\times6$   & 48.0  & 76.7 \\
    (d) D(1+2)D~(\textbf{ours}) & 3.2M  & $6.5\times6$   & 48.0  & 77.2 \\
    (e) D(2+1)D~(\textbf{ours}) & \textbf{3.2M}  & $\textbf{6.4}\times6$   & \textbf{49.0}  & \textbf{78.2} \\
    \midrule
    X3D-M~\cite{feichtenhofer2020x3d} & 3.3M & $6.1\times6$ & 46.4 &75.3 \\
    X3D-M~\cite{feichtenhofer2020x3d} w/ D(2+1)D & 3.2M & $5.8\times6$ & 47.4 & 75.9 \\
    \bottomrule
    \end{tabular}%
  }
\caption{\textbf{Comparison to different bottleneck architectures} on Something-Something V1.}
  \label{tab:dd}%
  \vspace{-0.3cm}
\end{table}%


\subsection{Implementation Details}

\noindent
\textbf{Training.} Our models are trained from scratch without using ImageNet~\cite{russakovsky2015imagenet} pretrained model unless specified.
For Something-Something~\cite{goyal2017something} dataset, we use segment-based input frame sampling~\cite{lin2019tsm}, which splits each video into $N$ segments and picks one frame to form a clip~($N$ frames) from each segment.
We note that thanks to the memory efficient VoV3D, our model can be trained with more input frames, \eg, from 16 to 32.
For Kinetics-400~\cite{kay2017kinetics}, we sample 16 frames with a temporal stride of 5 as \cite{feichtenhofer2020x3d}.
We apply the random cropping of $224\times224$ pixels from a clip and random horizontal flip with a shorter side randomly sampled in [256, 320] pixels~\cite{simonyan2014very,wang2018non,feichtenhofer2019slowfast,feichtenhofer2020x3d} for VoV3D-M and VoV3D-L models.
In case of Something-Something, it requires discriminating between directions, so the random flip is not applied.
Following~\cite{feichtenhofer2019slowfast,feichtenhofer2020x3d}, we use the same parameters for training Something-Something V1 \& V2: SGD optimizer, 100 epochs, mini-batch size 64~(8 clips per a GPU), initial learning rate 0.1, half-period cosine learning rate schedule~\cite{loshchilov2016sgdr}, linear warm-up strategy~\cite{goyal2017accurate}, and weight decay $5\times10^{-5}$.
Following~\cite{lin2019tsm,wu2020multigrid}, we also fine-tune VoV3D using Kinetics-400 pretrained model.
We use a linear warm-up~\cite{goyal2017accurate} for 2k iterations from 0.0001 and a weight decay of $5\times10^{-5}$.
We finetune the model for 50 epochs with a base learning rate of 0.05 decreased at 35 and 45 epoch by 0.1 and use sync bathcnorm.
For Kinetics-400, we use the same training parameters except for 256 epochs and mini-batch size 128.
We train all models using a 8-GPU machine and implementation is based on \texttt{PySlowFast}~\cite{fan2020pyslowfast}.

To compare VoV3D-M/L to the strong state-of-the-art X3D~\cite{feichtenhofer2020x3d}, we also train X3D-M/L having similar parameters and FLOPs with the same training protocols.
Note that for X3D-L, unlike origin X3D paper~\cite{feichtenhofer2020x3d}, we use the same spatial sample size [256, 320], not [356, 446].
The reason why we invest computation budget to more input frames ($\geq$16) is that the Something-Something dataset~\cite{goyal2017something} requires more temporal modeling than appearance information.

\begin{table}[t]
  \scalebox{0.96}{
    \begin{tabular}{lcccccc}
    \toprule
    \multicolumn{1}{c}{\multirow{2}[4]{*}{Model}} & \multicolumn{1}{c}{\multirow{2}[4]{*}{\#F}} & \multicolumn{1}{c}{\multirow{2}[4]{*}{GFLOPs}} & \multicolumn{2}{c}{From scratch} & \multicolumn{2}{c}{K-400 finetune} \\
\cmidrule{4-7}          &       &       & \multicolumn{1}{c}{Top-1} & \multicolumn{1}{c}{Top-5} & \multicolumn{1}{c}{Top-1} & \multicolumn{1}{c}{Top-5} \\
    \midrule
    X-M~\cite{feichtenhofer2020x3d} & 16    & 6.1$\times6$ & 46.4  & 75.3  & 51.2  & 78.9 \\
    \textbf{V-M} & 16    & 6.4$\times6$ & \textbf{49.0} & \textbf{78.2} & \textbf{52.4} & \textbf{80.3} \\
    X-M~\cite{feichtenhofer2020x3d} & 32    & 12.3$\times6$ & 48.9  & 77.6  & 51.5  & 79.6 \\
    \textbf{V-M} & 32    & 12.8$\times6$ & \textbf{50.1} & \textbf{79.2} & \textbf{53.3} & \textbf{81.2} \\
    \midrule
    X-L~\cite{feichtenhofer2020x3d} & 16    & 11.9$\times6$ & 47.1  & 76.5  & 50.8  & 79.3 \\
    \textbf{V-L} & 16    & 12.1$\times6$ & \textbf{49.6} & \textbf{78.1} & \textbf{53.4} & \textbf{81.4} \\
    X-L~\cite{feichtenhofer2020x3d} & 32    & 23.9$\times6$ & 48.4  & 77.8  & 52.6  & 81.2 \\
    \textbf{V-L} & 32    & 24.3$\times6$ & \textbf{50.7} & \textbf{78.8} & \textbf{54.7} & \textbf{82.0} \\
    \bottomrule
    \end{tabular}%
    }
  \caption{\textbf{Comparison to X3D on Something-Something V1.} \#F denotes the number of input frames. X and V denote X3D and VoV3D, respectively.
  For model parameters, X3D-M and VoV3D-M have 3.3M respectively and X3D-L and VoV3D-L have 5.6M and 5.8M, respectively.}
  \label{tab:x3d}%
  \vspace{-0.3cm}
\end{table}%

\noindent
\textbf{Inference.} 
Following common practice in~\cite{wang2018non,lin2019tsm,fan2020pyslowfast,feichtenhofer2020x3d}, we sample multiple clips per video (\textit{e.g.}, 10 for Kinetics and 2 for Something-Something).
We scale the shorter spatial side to 256 pixels and take 3 crops of 256$\times$256, as an approximation of fully-convolutional testing~\cite{wang2018non} called full resolution image testing in TSM~\cite{lin2019tsm}. 
Then, we average the softmax scores for prediction.

\begin{table*}[t]
  \centering
  \scalebox{0.97}{
    \begin{tabular}{lccccrcccc}
    \toprule
    \multicolumn{1}{c}{\multirow{2}[2]{*}{Model}} & \multicolumn{1}{c}{\multirow{2}[2]{*}{Backbone}} & \multicolumn{1}{c}{\multirow{2}[2]{*}{Pretrain}}& \multicolumn{1}{c}{\multirow{2}[2]{*}{Frame}} & \multicolumn{1}{c}{\multirow{2}[2]{*}{Param~(M)}} & \multicolumn{1}{c}{\multirow{2}[2]{*}{GFLOPs}} & \multicolumn{2}{c}{Something V1} & \multicolumn{2}{c}{SomethingV2} \\
          & &       &       &       &       & \multicolumn{1}{c}{Top-1} & \multicolumn{1}{c}{Top-5} & \multicolumn{1}{c}{Top-1} & \multicolumn{1}{c}{Top-5} \\
    \midrule
    TSM~\cite{lin2019tsm}                   &ResNet-50  & Kinetics & 16    & 24.3  & $33\times6$    & 47.2  & 77.0 & 63.0 & 88.1 \\
    TSM~\cite{lin2019tsm}                    &ResNet-101  & Kinetics & 8     & 24.3  & $65\times6$    & 48.7  & 77.2 & 63.2 & 88.2 \\
    TSM+TPN~\cite{yang2020temporal}           &ResNet-50  & ImageNet & 8     &  N/A  &   N/A          & 50.7  &  -    & 64.7  & - \\
    STM~\cite{jiang2019stm}                   &ResNet-50  & ImageNet & 16    &  N/A  & $67\times30$   & 50.7  & 80.4  & 64.2  & 89.8 \\
    TEA~\cite{li2020tea}                      &ResNet-50  & ImageNet & 8     & 24.4  & $35\times30$   & 51.7  & 80.5  &  -  &  -  \\
    TEA~\cite{li2020tea}                      &ResNet-50  & ImageNet & 16    & 24.4  & $70\times30$   & \textbf{52.3}  & \textbf{81.9}  & \textbf{65.1}  & \textbf{89.9} \\
    NL-I3D+GCN~\cite{wang2018videos}              &3D ResNet-50  & Kinetics & 32    &  N/A  & $303\times6$   & 46.1  & 76.8  &  -  & - \\
    SlowFast 16x8, R50~\cite{feichtenhofer2019slowfast} &- & Kinetics & 64    & 34.0 & $131.4\times6$ &   -   &   -   & 63.9 & 88.2 \\
    ip-CSN-152~\cite{tran2019video}               &- &   -      & 32    & 29.7  & $74.0\times10$ & 49.3  &   -   &   -   & - \\
    \midrule
    X3D-M~\cite{feichtenhofer2020x3d}             &- &   -      & 16    & 3.3   & $6.1\times6$   & 46.4 & 75.3 & 63.1 & 88.0 \\
    \textbf{VoV3D-M}                              &- &   -      & 16    & 3.2  & $6.4\times6$   & 49.0 & 78.2 & 63.6 & 88.6 \\
    \textbf{VoV3D-M}                              &- &   -      & 32    & 3.2  & $12.8\times6$  & 49.8 & 78.0 & 64.3 & 88.9 \\
    \textbf{VoV3D-M}                              &- & Kinetics & 32    & 3.2  & $12.8\times6$  & \textbf{53.2}  & \textbf{81.1} &  \textbf{65.8}  & \textbf{89.6}  \\
    \midrule
    X3D-L~\cite{feichtenhofer2020x3d}           &-   &    -     & 16    & 5.6   & $12.0\times6$   & 47.1 & 76.5 & 62.7 & 87.8 \\
    \textbf{VoV3D-L}                             &-  &    -     & 16    & 5.8   & $12.1\times6$   & 49.5  & 78.0 & 64.5 & 88.7 \\
    \textbf{VoV3D-L}                            &-   &    -     & 32    & 5.8   & $24.3\times6$  & 50.7 & 78.8 &65.9 & 89.6 \\
    \textbf{VoV3D-L}                            &-   & Kinetics & 32    & 5.8   & $24.3\times6$  & \textbf{54.7} & \textbf{82.0} & \textbf{67.4} & \textbf{90.5} \\
    \bottomrule
    \end{tabular}%
    }
  \caption{\textbf{Comparison with the state-of-the-art architectures on Something-Something V1\& V2 validation set.} Note that Something-Something dataset requires more temporal relationship than Kinetics-400~\cite{kay2017kinetics}~(appearance-oriented). For fair comparison, X3D and VoV3D are trained with the same training protocols on~\texttt{PySlowFast}~\cite{fan2020pyslowfast}.}
  \label{tab:ss}%
  \vspace{-0.2cm}
\end{table*}%

\subsection{Ablation study}
In order to verify the effectiveness of the proposed method in terms of temporal modeling and computational complexity, we conduct ablation studies on Something-Something V1~\cite{goyal2017something} that requires more temporal modeling ability~\cite{zhou2018temporal,xie2018rethinking,lin2019tsm} than Kinetics-400.

\noindent
\textbf{Component contributions.} We study the effect of the individual component of VoV3D and results are shown in Table~\ref{tab:ablation}.
We use X3D as a baseline and T-OSA without D(2+1)D consists of the same depthwise bottleneck as X3D.
T-OSA boosts performance by large margins in both M and L models, demonstrating the diverse temporal representation of T-OSA improves temporal modeling capability.
D(2+1)D also achieves higher accuracy, which suggests that the factorization of spatial and temporal features helps the network to optimize easily.

\noindent
\textbf{Comparison with the different bottleneck.} We compare the proposed depthwise spatiotemporal factorization module (\textit{i.e.}, D(2+1)D) with other architectures~\cite{Tran_2018_CVPR,feichtenhofer2019slowfast,tran2019video} in Table~\ref{tab:dd}.
We alternatively plug the bottleneck architectures into the T-OSA.
While R(2+1)D~\cite{Tran_2018_CVPR} reduces both parameters and GFLOPs with higher accuracy than the standard bottleneck~\cite{hara2018can} in (a), the depthwise bottleneck~\cite{feichtenhofer2020x3d,tran2019video} in (c) also significantly reduces the computations but obtains lower performance than R(2+1)D.
However, both D(1+2)D and D(2+1)D achieves better accuracy with less computation than dw-bottleneck in (c).
In particular, D(2+1)D outperforms all other architectures with a minimum computation and model size.
In addition, we also investigate the effect of D(2+1)D by replacing dw-bottleneck with D(2+1)D in X3D.
As a result, D(2+1)D improves 1\%p Top-1 accuracy gain while reducing model parameters and GFLOPs.

\noindent
\textbf{Comparison to X3D under various conditions.}
We compare VoV3D with X3D under the following conditions: the number of input frames (\#F in Table~\ref{tab:x3d}) and whether a backbone is pre-trained with Kinetics-400 or not.
We train VoV3D and X3D with 16 and 32 input frames from scratch or using Kinetics-400 pretraining.
Table~\ref{tab:x3d} summarizes the results.
We can find that using more frames boosts performance in both VoV3D and X3D and VoV3D consistently outperforms X3D.
This demonstrates that using more frames helps the networks to capture visual tempo variation and the ability of the proposed T-OSA to represent diverse temporal receptive fields enables VoV3D to yield better temporal modeling than X3D.

\subsection{Comparison to state-the-of-art}

\noindent
\textbf{Results on Something-Something.}
We validate the efficiency and effectiveness of the proposed VoV3D on Something-Something V1\&V2 requiring more temporal modeling ability than spatial appearance.
Table~\ref{tab:ss} shows the results and resource budgets (\textit{i.e.}, number of model parameters and GFLOPs) of other methods: temporal modeling based on 2D CNN methods~\cite{lin2019tsm,jiang2019stm,li2020tea} and 3D CNN architectures~\cite{wang2018videos,fan2020pyslowfast,yang2020temporal,fan2020pyslowfast,tran2019video,feichtenhofer2020x3d}.
First, under the same input frames (\textit{e.g.}, 16 frames), VoV3D-M/L consistently outperforms X3D-M/L with a comparable model budget on both Something V1\&V2. 
In particular, the performance gain of `L' models is bigger than `M' models.
This result demonstrates that stacking the proposed T-OSAs makes it better to model temporal dependency across frames.

Compared to the representative temporal modeling 2D CNN method, TSM~\cite{lin2019tsm} based ResNet-101, VoV3D-M with 16 frames achieves higher accuracy while it requires much fewer parameters~($8\times$) and GFLOPs~($10\times$), even without pretraining.
Furthermore, the performance of VoV3D-L with 32 frames pretrained on Kinetics-400 surpasses that of the best model among 2D CNN methods, TEA~\cite{li2020tea} by a large margin (2.4\% / 2.3\% @Top-1) on both SSv1/v2, while having about $14\times$ fewer computation.
These results break the prejudice that 3D CNN architectures require an expensive computation budget than 2D CNN.
We also note that VoV3D architecture alone shows sufficient performance and efficiency than the add-on style temporal modeling methods on top of 2D backbone networks~\cite{lin2019tsm,jiang2019stm,yang2020temporal,li2020tea}.
It shows that VoV3D can serve as a strong baseline for temporal modeling.

VoV3D is also superior to those 3D CNN-based temporal modeling methods, such as  SlowFast~\cite{feichtenhofer2019slowfast} and CSN~\cite{tran2019video}.
Even without Kinetics-pretraining, VoV3D-M with 32 frames achieves higher accuracy than SlowFast pretrained on Kinetics-400 with $11\times$ more model parameters.
It demonstrates that a 3D single network path is enough to model visual tempo variations.
Although CSN~\cite{tran2019video} contains the depthwise bottleneck architecture, its accuracy is lower than that of VoV3D-M.
This result shows that the proposed T-OSA plays an important role in temporal modeling. 

\begin{table}[t]
   \centering
  \scalebox{0.75}{
    \begin{tabular}{lcccccc}
    \toprule
    Method &Pre & \#F & P~(M) & GFLOPs & Top-1 & Top-5 \\
    \midrule
    I3D~\cite{carreira2017quo}   & IN &64 & 12    & $108\times N/A$ & 71.1  & 90.3 \\
    Nonlocal R50~\cite{wang2018non} & IN &32 & 35.3  & $282\times30$ & 76.5  & 92.6 \\
    TSM~\cite{lin2019tsm}  & IN  &16 & 24.3 & $65\times30$ & 74.7  & - \\
    STM~\cite{jiang2019stm}    & IN  &16 & N/A  & $67\times30$ & 73.7  & 91.6 \\
    TEA R50~\cite{li2020tea}  & IN &16 & 24.4  & $70\times30$ & 76.1  & 92.5 \\
    \midrule
    R(2+1)D~\cite{Tran_2018_CVPR} & -     &16 & 63.6  & $152\times115$ & 72.0    & 90.0 \\
    SlowFast 4$\times$16, R50~\cite{feichtenhofer2019slowfast} & -     &32 & 34.4  & $36.1\times30$ & 75.6  & 92.1 \\
    ip-CSN-152~\cite{tran2019video} & -     &32& 32.8  & $109\times30$ & \textbf{77.8}  & 92.8 \\
    X3D-M~\cite{feichtenhofer2020x3d} & -    &16 & 3.8   & $6.2\times30$ & 75.1   & 92.2 \\
    X3D-L~\cite{feichtenhofer2020x3d} & -    &16 & 6.1   & $9.1\times30$ & 76.1  & 92.6 \\
    \midrule
    VoV3D-M & -     &16& 3.7   & $6.4\times30$ & 74.7 & 92.1 \\
    VoV3D-L & -     &16& 6.2   & $9.3\times30$ & 76.3 & \textbf{92.9} \\
    \bottomrule
    \end{tabular}%
  }
  \caption{\textbf{Comparison with the state-of-the-art architectures on Kinetics-400.} IN denotes ImageNet pretraining. Note that both VoV3D and X3D are trained with the same training protocols on the same environment such as GPU server, training set, and scale size~[256, 320] and implemented on \texttt{PySlowFast}~\cite{fan2020pyslowfast}.
  }
  \label{tab:addlabel}%
\end{table}%

\noindent
\textbf{Results on Kinetics-400.}
We also compare VoV3D to other state-of-the-art methods on Kinetics-400.
VoV3D-L achieves 76.3\%/92.9\% Top-1/5 accuracy, and it shows better performance than the state-of-the-art temporal modeling 2D method, TEA~\cite{li2020tea}, even without ImageNet pretraining.
VoV3D-L also surpasses 3D temporal modeling methods, SlowFast~\cite{feichtenhofer2019slowfast} $4\times16$ based on ResNet-50 while having about $5\times$ and $4\times$ fewer model parameters and FLOPs, respectively.
Compared to ip-CSN-152~\cite{tran2019video} as an efficient 3D CNN, VoV3D-L shows slightly lower Top-1 accuracy, but it achieves higher Top-5 accuracy with much less model capacity.
While VoV3D-M shows comparable accuracy with X3D-M, VoV3D-L achieves higher  Top-1/Top-5 accuracy. 

\begin{figure}[t]
\centering
  \scalebox{0.47}{
  \includegraphics[width=\textwidth]{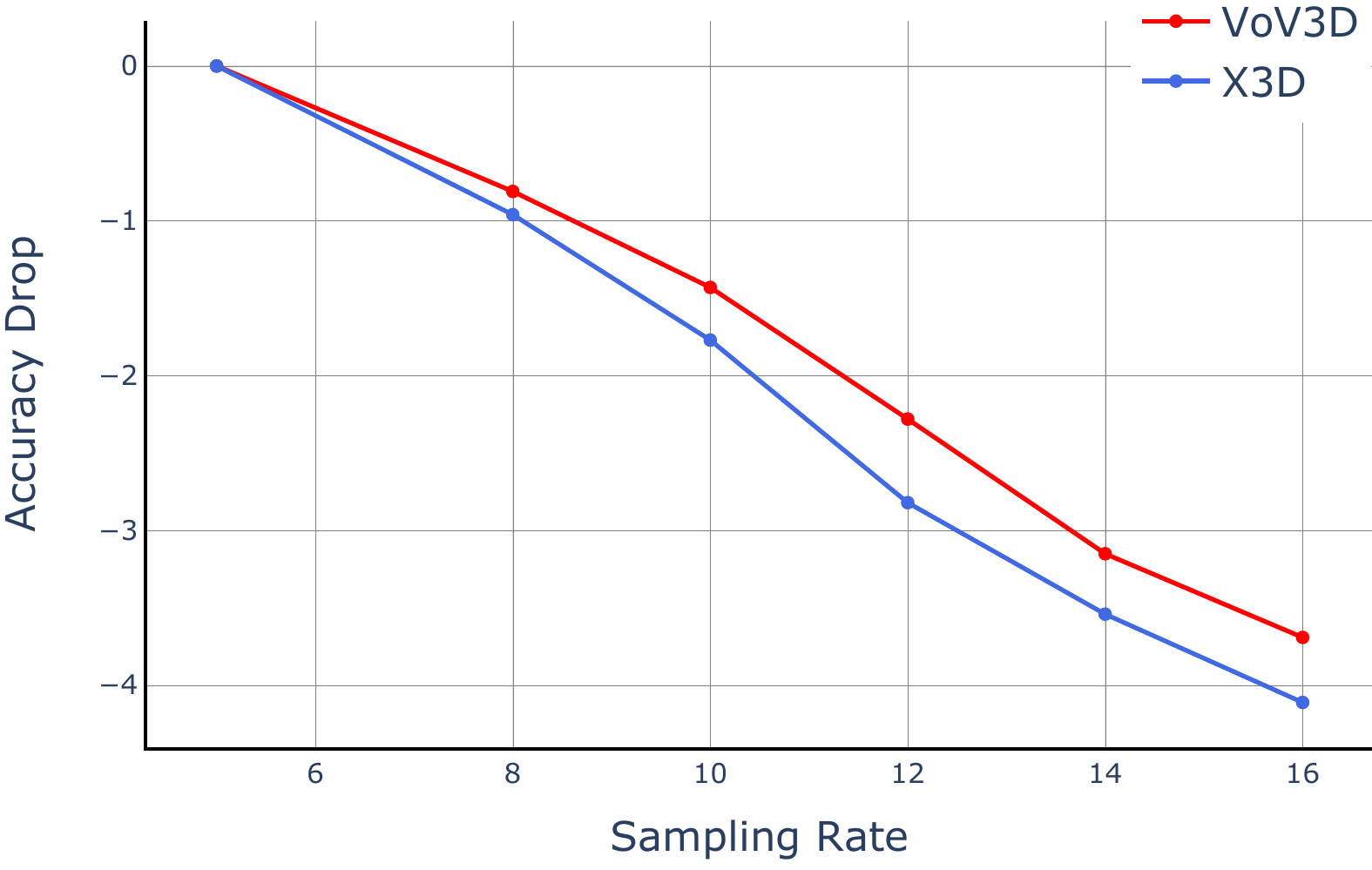} 
  }
  \caption{\textbf{Robustness to temporal variation.} Changing sampling rates~(or temporal stride) induces temporal variation. Compared to X3D, VoV3D is more robust to temporal variation due to its temporal modeling ability of T-OSA.
  }
\label{fig:robust}
\end{figure}

\section{Robustness Analysis to Temporal Variation}
Inspired by TPN~\cite{yang2020temporal}, we investigate the robustness to temporal variation of VoV3D and X3D. 
VoV3D-M and X3D-M are trained with the same sampling rate~(temporal stride $\tau$) of 5 on Kinetics-400 with 16 frames as input.
At the test phase, we measure the top-1 accuracy drop depending on the change of the sampling rate~(\eg, $\tau \in \{5,8,10, 12, 14, 16\}$) used for adjusting the visual tempo of a given action instance. The accuracy drop is used for measuring the robustness to temporal variations.
Fig.~\ref{fig:robust} shows the accuracy curves of varying visual tempos for VoV3D and X3D.
When changing the sampling rate, VoV3D shows less accuracy drop than X3D, which supports the fact that VoV3D is more robust to temporal variation and thus has a better ability to model temporal relationship across frames than X3D.

\section{Conclusion}
We have proposed an efficient and effective temporal modeling 3D architecture, called VoV3D, that consists of Temporal One-Shot Aggregation~(T-OSA) and depthwise spatiotemporal factorized module, D(2+1)D.
The T-OSA is able to effectively model various visual tempos by aggregating features having different temporal receptive fields.
The D(2+1D) module decomposes 3D depthwise convolution into a spatial and temporal depthwise convolution, which makes the proposed VoV3D significantly lightweight and efficient while improving accuracy.
Thanks to T-OSA and D(2+1)D, our VoV3D outperforms the state-of-the-art 2D efficient CNN as well as 3D CNN methods in terms of temporal modeling, with lower computational complexity. 
We hope that it can serve as a strong baseline for video action recognition.

\section{Acknowledgement}
This work was supported by Institute of Information \& Communications Technology Planning \& Evaluation (IITP) grant funded by the Korea government (MSIT) (No. 2020-0-00004, Development of Previsional Intelligence based on Long-term Visual Memory Network and No.B0101-15-0266, Development of High Performance Visual BigData Discovery Platform for Large-Scale Realtime Data Analysis)

{\small
\bibliographystyle{ieee_fullname}
\bibliography{ref}
}

\end{document}